\documentclass[letterpaper, 10 pt, conference]{ieeeconf}  

\IEEEoverridecommandlockouts                              

\usepackage{algorithm} 
\usepackage{algorithmic} 

\usepackage{subfigure}
\usepackage{multirow}
\usepackage[T1]{fontenc}
\usepackage{graphicx}
\usepackage{amsmath}

\title{\LARGE \bf
Group Feature Learning and Domain Adversarial Neural Network for aMCI Diagnosis System Based on EEG
}

\author{Chen-Chen Fan$^{*,1,2}$, Haiqun Xie$^{*,5}$, Liang Peng$^{1}$, Hongjun Yang$^{1}$, Zhen-Liang Ni$^{1,2}$, Guan’an Wang$^{1,2}$, \\Yan-Jie Zhou$^{1,2}$, Sheng Chen$^{1,2}$, Zhijie Fang$^{1,2}$, Shuyun Huang$^{5}$, Zeng-Guang Hou$^{1,2,3,4}$, \textit{Fellow}, \textit{IEEE}
\thanks{$^{1}$State Key Laboratory of Management and Control for Complex Systems, Institute of Automation, Chinese Academy of Sciences, Beijing 100190, China}%
\thanks{$^{2}$University of Chinese Academy of Sciences, Beijing 100049, China}%
\thanks{$^{3}$CAS Center for Excellence in Brain Science and Intelligence Technology, Beijing 100190, China}%
\thanks{$^{4}$Joint Laboratory of Intelligence Science and Technology, Institute of Systems Engineering, Macau University of Science and Technology, China}%
\thanks{$^{5}$Department of Neurology, First People’s Hospital of Foshan, Foshan, Guangdong 528000, China. Corresponding author: Zeng-Guang Hou. email:{\tt\small \{zengguang.hou,fanchenchen2018\}@ia.ac.cn}}%
\thanks{$^{*}$These authors contributed equally to this work}%
}

\begin{document}

\maketitle
\thispagestyle{empty}
\pagestyle{empty}

\begin{abstract}
Medical diagnostic robot systems have been paid more and more attention due to its objectivity and accuracy. 
The diagnosis of mild cognitive impairment (MCI) is considered an effective means to prevent Alzheimer's disease (AD).
Doctors diagnose MCI based on various clinical examinations, which are expensive and the diagnosis results rely on the knowledge of doctors.
Therefore, it is necessary to develop a robot diagnostic system to eliminate the influence of human factors and obtain a higher accuracy rate.
In this paper, we propose a novel Group Feature Domain Adversarial Neural Network (GF-DANN) for amnestic MCI (aMCI) diagnosis, which involves two important modules.
A Group Feature Extraction (GFE) module is proposed to reduce individual differences by learning group-level features through adversarial learning. 
A Dual Branch Domain Adaptation (DBDA) module is carefully designed to reduce the distribution difference between the source and target domain in a domain adaption way.
On three types of data set, GF-DANN achieves the best accuracy compared with classic machine learning and deep learning methods. On the DMS data set, GF-DANN has obtained an accuracy rate of 89.47\%, and the sensitivity and specificity are 90\% and 89\%. In addition, by comparing three EEG data collection paradigms, our results demonstrate that the DMS paradigm has the potential to build an aMCI diagnose robot system.

\end{abstract}

\section{INTRODUCTION}

Machine learning method can help the construction of more effective robots in the medical field \cite{zhou2020multilayer,zhou2019qualitative}, and has been widely used in disease diagnosis systems.
The World Alzheimer Report 2015 \cite{Report2015} points out that there are around 46 million people with Alzheimer's disease (AD) worldwide. Mild cognitive impairment (MCI) is considered as an intermediate transition state between normal aging and AD, \textit{i.e.} the early stage of AD.
MCI patients convert to AD at a rate of 10\% to 15\% per year, which is ten times that of normal elderly people \cite{Petersen2000Mild}. Amnestic mild cognitive impairment (aMCI) is a subtype of MCI, and memory impairment is the central and earliest symptom. Since there is no widely used specific drug for AD, early prevention of AD is critical. People with aMCI have a high risk of developing AD \cite{petersen2004mild}, so the diagnosis of aMCI plays a significant role in the early recognition and early intervention of AD.
However, the diagnosis of aMCI requires a variety of expensive medical examinations (e.g., MRI, PET) and the doctors of the high level. Due to the uneven distribution of medical resources, some people living in remote areas cannot timely screen for aMCI, resulting in a greatly increased risk of AD.
Therefore, the construction of a medical robot diagnostic system for aMCI diagnosis will help to improve the medical condition in remote areas.

As a widely used electrophysiological examination method, EEG can record the brain's spontaneous biological potential and reflect the brain's activity. Its advantages are higher time resolution, relatively low cost, and portable equipment. However, there are considerable differences in EEG signals among different individuals, which brings significant challenges to the diagnosis of aMCI.
Disease diagnosis based on EEG signals faces several challenges. The first one is robust feature extraction. In the same group, the data distributions among individuals are very different. 
It might force the neural network tend to remember individual features 
instead of general and key information of aMCI. 

The second one is domain bias. Besides the differences among individuals, the data distribution of the same person is also affected by the temporal-spatial factors. For example, morning (after having a good sleeping) and a quiet place contribute to EEG of high quality, but night (easy to be tired) and a noisy place affect EEG quality a lot. Thus there is no guarantee that the train and test data have the same or similar data distribution. The different distribution of train and test data makes the model performs worse on test data.

Recently, some studies have tried to diagnose MCI or aMCI. 
Most of these methods using various machine learning algorithms to build classifier \cite{Lehmann2007Application}. 
The regional spectral features of EEG and features based on complexity were extracted and classified by SVM \cite{Joseph2014Spectral}.  

Neural fuzzy k-nearest neighbor classifier was proposed for EEG classification \cite{Kashefpoor2016Automatic}.
Relative power (RP) was classified as a feature input neural network to diagnose MCI \cite{8512231}. 
A deep machine learning method based on convolutional neural network and autoencoder MLP was proposed to learn features \cite{7740576}.
Implicit compression time function (IFAST) compressed closed-eye resting EEG data into spatial invariants of instantaneous voltage distribution, and employing MLP for the classification task \cite{ROSSINI20081534}.

Most of these methods do not consider the difference in data distribution among the train and test data. Specifically, traditional supervised learning methods assume that the train and test data have the same or similar distribution, thus the model trained by the train set can be applied to predict the label of the test set. However, there are considerable differences in EEG signals among different individuals, and the train and test set cannot satisfy the assumption of similar distribution, thus limiting the performance of these methods.

To extract robust features related to aMCI, we propose a group feature extraction framework GFE (Figure \ref{fig:GF-DANN}). All train data is divided into two groups according to labels (aMCI and HC: healthy control), and each group of data contains multiple individuals (each individual has multiple samples). For the samples in each group, a feature extractor is used to generate a feature representation. The individual discriminator is used to determine which individual the sample comes from. Through the adversarial learning, the feature extractor generates a feature representation that confuses the individual discriminator. These group-level features reduce the difference among different individuals in the same group and emphasize the corresponding group attributes. Thus group-level features have more potential to be truly relevant to the disease.

To alleviate the domain bias, 
we propose a Dual Branch Domain Adaptation architecture (DBDA) as in Figure \ref{fig:GF-DANN}, 
which enables a more general assumption that train and test data may not maintain the same distribution.
DBDA reduces the difference in the marginal distribution among the train and test data through adversarial learning on the premise of the unknown test data label. 
Specifically, through the adversarial learning of feature extractor and domain discriminator, the feature extractor transforms the original data into a new feature space, where the marginal distribution of the train and test data is more similar.

Based on the above analysis, a group feature domain adversarial neural
network (GF-DANN) framework (Figure \ref{fig:GF-DANN}) is proposed, which combines GFE and DBDA for multi-task learning.
GF-DANN extracts more robust group features and uses adversarial learning to solve the domain adaptation problem.

The contributions of this work can be concluded as follows:

\begin{itemize}
	\item We design the GFE module to learn group features by adversarial learning.
	\item We introduce DBDA module to reduce the distribution difference between the source and target domains.
	\item The proposed method has achieved significantly better results than classic deep learning and machine learning algorithms in all three data sets.
\end{itemize}

\section{Overview of GF-DANN Architechture}

\begin{figure*}[thpb]
	\centering
	\framebox{\parbox{5.6in}{\includegraphics[scale=0.46]{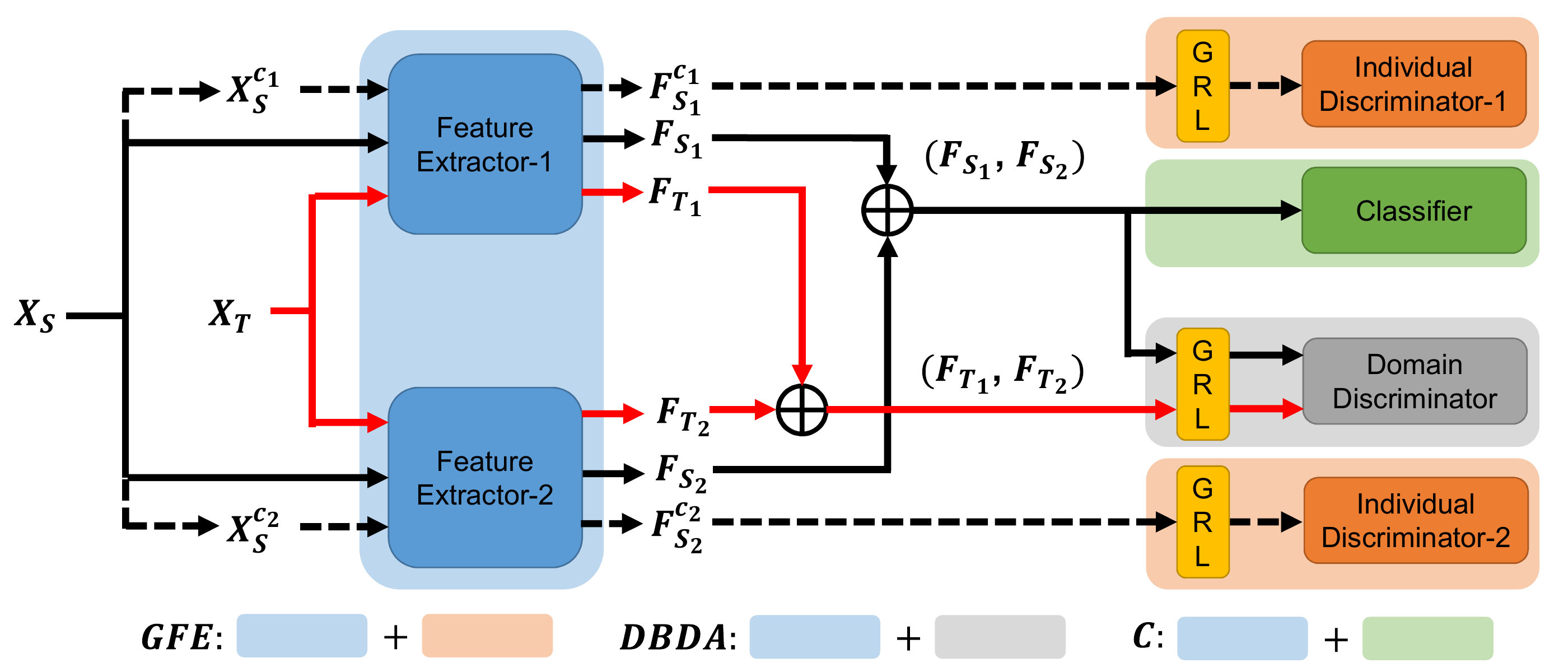}}}
	\caption{The framework of GF-DANN. It contains three modules: Group Feature Extraction (GFE), Dual Branch Domain Adaptation (DBDA), and Classification (C). The solid black line represents the source domain data, the solid red line is the target domain data, and the black dashed line denotes a specific category of train data.}
	\label{fig:GF-DANN}
\end{figure*}

The architechture of GF-DANN is shown in Figure \ref{fig:GF-DANN}. 
It contains three modules: Group Feature Extraction, Dual Branch Domain Adaptation, and Classification, which correspond to three learning tasks respectively. Through multi-task training, GF-DANN extracts more robust group features and reduces the distribution difference between the source domain and the target domain.
The detailed mathematical description of GF-DANN is given below.

\begin{figure}[tbp]
	\centering
	\includegraphics[width=1\linewidth]{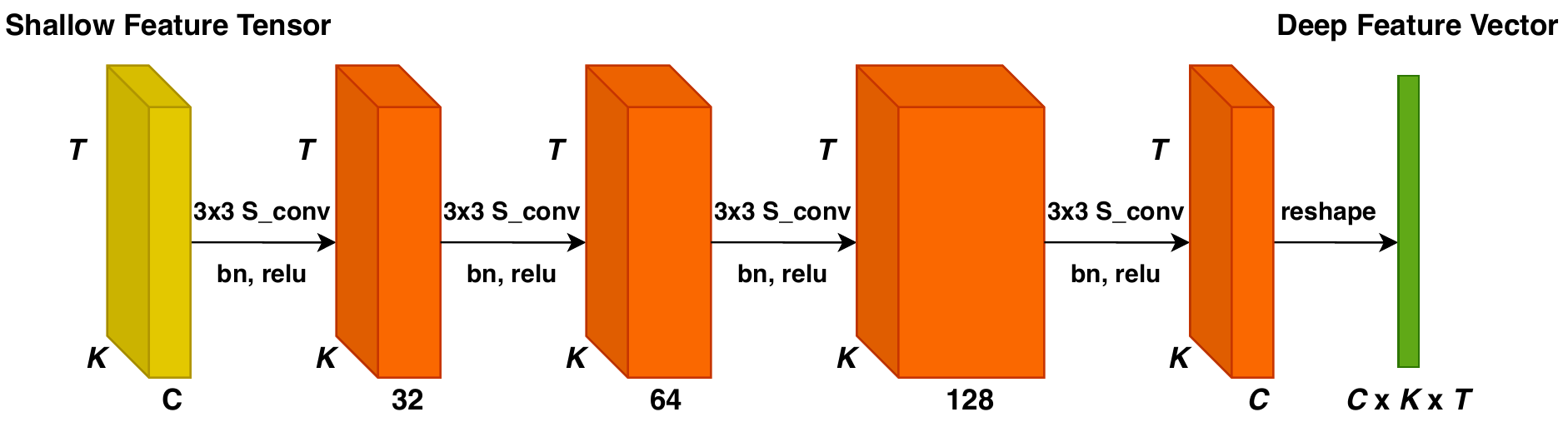}
	\caption{The feature extractor of GF-DANN.}
	\label{fig:Feature_Extractor}
\end{figure}

\subsection{Group Feature Extraction Module}

EEG data has individual differences, even in the same group.
To extract robust features related to aMCI, GFE module learns the common features in each group through two pairs of individual discriminators and feature extractors to eliminate the influence of individual differences. It includes two feature extractors, two individual discriminators, and two gradient reverse layers. 

\textbf{Feature Extractor:} The input data (shallow feature tensor) is defined as $x_{(u, v)}^{c_{i}} \in R^{C \times K \times T}$, where $c_{i} \in\left\{c_{1}, c_{2}\right\}$ indicates that the sample is from the aMCI group or HC group. The individual label set is defined as $I_{1}=\{1, \cdots, \mathrm{i}, \cdots, m\}$ and $I_{2}=\{1, \cdots, \mathrm{j}, \cdots, n\}$, where $m$ ($n$) is the total number of aMCI (HC) group subjects. $u \in I_{1}$ (if $c_{i} = c_{1}$) or $u \in I_{2}$ (if $c_{i} = c_{2}$) is the individual label in the corresponding group, and $v$ indicates the individual sample number. Therefore, $x_{(u, v)}^{c_{i}}$ represents the $v$-th sample of the individual with the individual number $u$ in the group $c_{i}$. $C$ represents the dimension of the feature, $K$ represents the number of frequency bands, and $T$ represents the number of time bands. The acquisition of input data is described in detail in section \ref{section:Shallow Feature Extraction}. 

The structure of the feature extractor is shown in Figure \ref{fig:Feature_Extractor}. As the input data is not an image, pooling layer is not used in order not to lose the feature information. To reduce the amount of parameters and computation cost, the input data pass through three $3 \times 3$ depthwise separable convolution \cite{howard2017mobilenets} with bn and relu activation layers to extract features. Then the $1 \times 1$ convolution is used to reduce the number of channels and finally stretch the extracted features to one dimension to obtain the group feature vector $F$.

Define $X_{R}$ to represent all samples. According to the samples from the source domain and the target domain, $X_{R} = X_{S} \cup X_{T}$ can be obtained, where $X_{S}$ represents the source domain data set and $X_{T}$ represents the target domain data set. Define the set $M=\left\{p_{1}, p_{2}, \cdots, p_{i}, \cdots, p_{m}\right\}$ and $N=\left\{q_{1}, q_{2}, \cdots, q_{j}, \cdots, q_{n}\right\}$, where $m$ ($n$) is the number of subjects in group $c_{1}$ ($c_{2}$) and $p_{i}$ ($q_{j}$) represents the number of samples owned by subject with individual label $i$ ($j$) from the group $c_{1}$ ($c_{2}$). According to the group label of the train set, $X_{S}$ can be expressed as:
\begin{footnotesize} 
	\begin{equation}\label{key1}
	X_{S}=X_{S}^{c_{1}} \cup X_{S}^{c_{2}}
	\end{equation}
	\begin{equation}\label{key2}
	X_{S}^{c_{1}}=X_{1}^{c_{1}} \cup X_{2}^{c_{1}} \cdots \cup X_{i}^{c_{1}} \cdots \cup X_{m}^{c_{1}}
	\end{equation}
	\begin{equation}\label{key3}
	X_{S}^{c_{2}}=X_{1}^{c_{2}} \cup X_{2}^{c_{2}} \cdots \cup X_{j}^{c_{2}} \cdots \cup X_{n}^{c_{2}}
	\end{equation}
\end{footnotesize}  

\noindent where $X_{i}^{c_{1}}=\left\{x_{(i, 1)}^{c_{1}}, x_{(i, 2)}^{c_{1}}, \cdots, x_{\left(i, p_{i}\right)}^{c_{1}}\right\}$,$1 \leq i \leq m$,  $X_{j}^{c_{2}}=\left\{x_{(j, 1)}^{c_{2}}, x_{(j, 2)}^{c_{2}}, \cdots, x_{\left(j, q_{j}\right)}^{c_{2}}\right\}$, $1 \leq j \leq n$.

The features extracted by the feature extractor can be formulated as:
\begin{footnotesize} 
	\begin{equation}\label{key4}
	F_{S_{k}}^{c_{k}}=\left\{\xi_{k}(x) | x \in X_{S}^{c_{k}}\right\}
	\end{equation}
	\begin{equation}\label{key5}
	F_{S_{k}}=\left\{\xi_{k}(x) | x \in X_{S}\right\}
	\end{equation}
	\begin{equation}\label{key6}
	F_{T_{k}}=\left\{\xi_{k}(x) | x \in X_{T}\right\}
	\end{equation}
\end{footnotesize} 

\noindent where $k \in \left\{1,2\right\}$, $\xi_{k}: R^{C \times K \times T} \rightarrow R^{d}$ denotes the function map of feature extractor-$k$, $d$ is the length of the feature vector.


\textbf{Individual Discriminator:} Individual discriminators 1 and 2 are trained by individual label sets $I_{1}$ and $I_{2}$, respectively. 
GRL is a module proposed by \cite{ganin2016domain-adversarial}, which keeps the input unchanged during forwarding propagation, and multiplies the negative scalar during reverse propagation to reverse the gradient, which can maximize the loss of the discriminator.
The individual discriminator strives to distinguish which subject the input data come from, while the feature extractor updates the parameters by maximizing the loss of individual discriminator and strives to generate a data representation to confuse the individual discriminator. The loss function of the individual discriminator is defined as follows:
\begin{footnotesize} 
	\begin{equation}
	L_{d_{1}}\left(X_{S}^{c_{1}} ; \theta_{f_{1}}, \theta_{d_{1}}\right)=\ell_{d_{1}}\left(\psi_{d_{1}}\left(\xi_{1}\left(X_{S}^{c_{1}} ; \theta_{f_{1}}\right) ; \theta_{d_{1}}\right), Y_{d_{1}}\right)
	\end{equation}
	\begin{equation}
	L_{d_{2}}\left(X_{S}^{c_{2}} ; \theta_{f_{2}}, \theta_{d_{2}}\right)=\ell_{d_{2}}\left(\psi_{d_{2}}\left(\xi_{2}\left(X_{S}^{c_{2}} ; \theta_{f_{2}}\right) ; \theta_{d_{2}}\right), Y_{d_{2}}\right)
	\end{equation}
\end{footnotesize} 

\noindent where $\ell_{d_{1}}, \ell_{d_{2}}$ denote loss functions, $\psi_{d_{1}}, \psi_{d_{2}}$ are individual label classifiers, $Y_{d_{1}} \in I_{1}, Y_{d_{2}} \in I_{2}$ represent individual labels, $\theta_{f_{1}}, \theta_{f_{2}}, \theta_{d_{1}}, \theta_{d_{2}}$ are the parameters of the two feature extractors and two individual discriminators.

	\subsection{Dual Branch Domain Adaptation Module}
There are differences among the source domain and target domain data of EEG signals. To reduce this difference, DBDA module extracts features through a dual-branch feature extractor and conducts adversarial training with the domain discriminator.
DBDA includes two feature extractors (shared with GFE), two individual discriminators, and a gradient reverse layer. 

\textbf{Domain Discriminator:} The domain discriminator is trained by the domain label set $D=\{0,1\}$, 
Where $0$ and $1$ represent data from the source domain and target domain, respectively. 
Domain discriminator narrows the feature distribution gap between the source and target domains. 
Specifically, the domain discriminator determines which domain the input data come from. The two feature extractors update their parameters by maximizing the loss of the domain discriminator and strive to generate data representations to confuse the judgment of the domain discriminator.
The loss function of the domain discriminator is defined as follows:
\begin{footnotesize}
	\begin{equation}
	L_{d_{3}}\left(X_{R} ; \theta_{f_{1}}, \theta_{f_{2}}, \theta_{d_{3}}\right)=\ell_{d_{3}}\left(\tilde{Y}_{d_{3}}, Y_{d_{3}}\right)	
	\end{equation}
	\begin{equation}
	\tilde{Y}_{d_{3}}=\psi_{d_{3}}\left(\xi_{1}\left(X_{R} ; \theta_{f_{1}}\right),\xi_{2}\left(X_{R} ; \theta_{f_{2}}\right) ; \theta_{d_{3}}\right)
	\end{equation}
\end{footnotesize}

\noindent where $\ell_{d_{3}}$ denotes loss function, $\tilde{Y}_{d_{3}} \in D$ represents the prediction result of $\psi_{d_{3}}$, $Y_{d_{3}} \in D$ denotes domain label, $\psi_{d_{3}}$ is domain label classifier, $\theta_{d_{3}}$ represents the parameters of domain discriminator.
\begin{algorithm}[t] 
	\small
	\caption{Optimization of GF-DANN} 
	\label{alg1} 
	\begin{algorithmic}[1] 
		\REQUIRE 
		\qquad 
		
		Individual discriminator train data: $X_{S}^{c_{1}}, X_{S}^{c_{2}}$;
		
		Individual discriminator train label: $Y_{d_{1}}, Y_{d_{2}}$;
		
		Domain discriminator train data:
		$X_{R}$;
		
		Domain discriminator train label:
		$Y_{d_{3}}$;
		
		Classifier train data:
		$X_{S}$;
		
		Classifier train label:
		$Y_{c}$;
		
		Initial learning rate: $\alpha_{d_{1}}, \alpha_{d_{2}}, \alpha_{d_{3}}, \alpha_{c}$
		
		Discriminator and classifier train epochs: $n_{d}$, $n_{c}$
		
		\ENSURE $\hat{\theta}_{f_{1}}, \hat{\theta}_{f_{2}}, \hat{\theta}_{c}, \hat{\theta}_{d_{1}}, \hat{\theta}_{d_{2}}, \hat{\theta}_{d_{3}}$
		\FOR{$i=1$ to $n_{d}$} 
		\STATE 
		Input $X_{S}^{c_{1}}$ and $Y_{d_{1}}$ to update parameters $\theta_{f_{1}}, \theta_{d_{1}}$:
		
		$\theta_{f_{1}} \leftarrow \theta_{f_{1}}+\alpha_{d_{1}} \frac{\partial L_{d_{1}}}{\partial \theta_{f_{1}}}, \theta_{d_{1}} \leftarrow \theta_{d_{1}}-\alpha_{d_{1}} \frac{\partial L_{d_{1}}}{\partial \theta_{d_{1}}}$;
		\STATE 
		Input $X_{S}^{c_{2}}$ and $Y_{d_{2}}$ to update parameters $\theta_{f_{2}}, \theta_{d_{2}}$:
		
		$\theta_{f_{2}} \leftarrow \theta_{f_{2}}+\alpha_{d_{2}} \frac{\partial L_{d_{2}}}{\partial \theta_{f_{2}}}, \theta_{d_{2}} \leftarrow \theta_{d_{2}}-\alpha_{d_{2}} \frac{\partial L_{d_{2}}}{\partial \theta_{d_{2}}}$;
		\STATE 
		Input $X_{R}$ and $Y_{d_{3}}$ to update parameters $\theta_{f_{1}}, \theta_{f_{2}}, \theta_{d_{3}}$:
		
		$\theta_{f_{1}} \leftarrow \theta_{f_{1}}+\alpha_{d_{3}} \frac{\partial L_{d_{3}}}{\partial \theta_{f_{1}}},
		\theta_{f_{2}} \leftarrow \theta_{f_{2}}+\alpha_{d_{3}} \frac{\partial L_{d_{3}}}{\partial \theta_{f_{2}}}$, 
		
		$\theta_{d_{3}} \leftarrow \theta_{d_{3}}-\alpha_{d_{3}} \frac{\partial L_{d_{3}}}{\partial \theta_{d_{3}}}$;
		
		\ENDFOR 
		\FOR{$j=1$ to $n_{c}$} 
		\STATE 
		Input $X_{S}$ and $Y_{c}$ to update parameters $\theta_{f_{1}}, \theta_{f_{2}}, \theta_{c}$:
		
		$\theta_{f_{1}} \leftarrow \theta_{f_{1}}-\alpha_{c} \frac{\partial L_{c}}{\partial \theta_{f_{1}}},
		\theta_{f_{2}} \leftarrow \theta_{f_{2}}-\alpha_{c} \frac{\partial L_{c}}{\partial \theta_{f_{2}}}$, 
		
		$\theta_{c} \leftarrow \theta_{c}-\alpha_{c} \frac{\partial L_{c}}{\partial \theta_{c}}$;
		\ENDFOR 
		\RETURN $\hat{\theta}_{f_{1}}, \hat{\theta}_{f_{2}}, \hat{\theta}_{c}, \hat{\theta}_{d_{1}}, \hat{\theta}_{d_{2}}, \hat{\theta}_{d_{3}}$.
		
	\end{algorithmic} 
\end{algorithm}
\subsection{Classification Module}

To diagnose aMCI, the Classifier shares the feature extractor with GFE and DBDA to use the learned features. Voting Diagnosis Framework is proposed to give the diagnosis result.


\textbf{Classifier:} Fully connected layer and softmax are used to obtain the classification results, which can be expressed as:
\begin{footnotesize}
	\begin{equation}
	f_{S}=\left[f_{S_{1}}^{T}, f_{S_{2}}^{T}\right]^{T}
	\end{equation}
	\begin{equation}
	\tilde{y}=\eta\left(A * f_{S}+b\right)
	\end{equation}
\end{footnotesize}

\noindent where $f_{S_{1}} \in F_{S_{1}}$, $f_{S_{2}} \in F_{S_{2}}$, $f_{S} \in R^{2d}$, $A \in R^{2 \times 2d}$, $b \in R^{2}$, $\eta$ denote softmax function. The loss function of classifier can be expressed as:
\begin{footnotesize}
	\begin{equation}
	L_{c}\left(X_{S} ; \theta_{f_{1}}, \theta_{f_{2}}, \theta_{c}\right)=\ell_{c}\left(\tilde{Y}_{c}, Y_{c}\right)
	\end{equation}
	\begin{equation}
	\tilde{Y}_{c}=\psi_{c}\left(\xi_{1}\left(X_{S} ; \theta_{f_{1}}\right), \xi_{2}\left(X_{S} ; \theta_{f_{2}}\right) ; \theta_{c}\right)
	\end{equation}
\end{footnotesize}

\noindent where $\ell_{c}$ denotes loss function, $\hat{Y}_{c} \in \left\{0,1\right\}$ represents the prediction result of $\psi_{c}$, $Y_{c} \in \left\{0,1\right\}$ denotes true group label, $\psi_{c}$ is classifier, $\theta_{c}$ represents the parameters of classifier.


\textbf{Voting Diagnosis Framework:} Since each subject has multiple samples, in order to give the diagnosis results of the subject, we propose the voting diagnosis framework 
. In the testing phase, all $k$ samples of a single subject obtain the corresponding sample-level results $\tilde{y}_{i} \in \{0,1\},i=1,\cdots,k$ through GF-DANN.
\begin{footnotesize} 
	\begin{equation}\label{key15}
	\tilde{y}_{\text{subject}}=\left\{\begin{array}{c}
	1, \text{if}  \sum_{i=1}^{k} \tilde{y}_{i}>\frac{k}{2} \\
	0,  \text{if} \sum_{i=1}^{k} \tilde{y}_{i}<\frac{k}{2} \\
	\text{refuse}, \text{if}  \sum_{i=1}^{k} \tilde{y}_{i}=\frac{k}{2}
	\end{array}\right.
	\end{equation}
\end{footnotesize} 

\section{Optimization of GF-DANN}

The overall loss function is defined as:
\begin{footnotesize} 
	\begin{equation}\label{key16}
	L\left(\mathbf{X}_{\mathcal{R}} ; \theta_{f_{1}}, \theta_{f_{2}}, \theta_{c}, \theta_{d_{1}}, \theta_{d_{2}}, \theta_{d_{3}}\right) = L_{c} - L_{d_{1}} - L_{d_{2}} - L_{d_{3}}
	\end{equation}
\end{footnotesize} 

The optimization of GF-DANN can be described as the following minimax problem: 
\begin{footnotesize} 
	\begin{equation}\label{key17}
	\min _{\theta_{f_{1}}, \theta_{f_{2}}, \theta_{c}} \max _{\theta_{d_{1}}, \theta_{d_{2}}, \theta_{d_{3}}} L\left(\mathbf{X}_{\mathcal{R}} ; \theta_{f_{1}}, \theta_{f_{2}}, \theta_{c}, \theta_{d_{1}}, \theta_{d_{2}}, \theta_{d_{3}}\right)
	\end{equation}
\end{footnotesize} 

The optimization algorithm of GF-DANN is shown in algorithm \ref{alg1}. The optimization problem can be described as a minimax problem:
\begin{footnotesize} 
	\begin{equation}\label{key18}
	\hat{\theta}_{f_{1}}, \hat{\theta}_{d_{1}}=\arg \min _{\theta_{d_{1}}} \max _{\theta_{f_{1}}} L_{d_{1}}\left(X_{S}^{c_{1}} ; \theta_{f_{1}}, \theta_{d_{1}}\right)
	\end{equation}
	\begin{equation}\label{key19}
	\hat{\theta}_{f_{2}}, \hat{\theta}_{d_{2}}=\arg \min _{\theta_{d_{2}}} \max _{\theta_{f_{2}}} L_{d_{2}}\left(X_{S}^{c_{2}} ; \theta_{f_{2}}, \theta_{d_{2}}\right)
	\end{equation}
	\begin{equation}\label{key20}
	\hat{\theta}_{f_{1}}, \hat{\theta}_{f_{2}}, \hat{\theta}_{d_{3}}=\arg \min _{\theta_{d_{3}}} \max _{\theta_{f_{1}}, \theta_{f_{2}}} L_{d_{3}}\left(X_{R} ; \theta_{f_{1}}, \theta_{f_{2}}, \theta_{d_{3}}\right)
	\end{equation}
	\begin{equation}\label{key21}
	\hat{\theta}_{f_{1}}, \hat{\theta}_{f_{2}}, \hat{\theta}_{c}=\arg \min _{\theta_{f}, \theta_{2}, \theta_{c}} L_{c}\left(X_{S} ; \theta_{f_{1}}, \theta_{f_{2}}, \theta_{c}\right)
	\end{equation}
\end{footnotesize}

To solve the minimax optimization problem of Eq.(\ref{key18}-\ref{key20}), we use the gradient reverse layer (GRL) to transform the problem into a minimization problem. 
Adam optimizer is used to minimize $L_{d_{1}} \left( L_{d_{2}}, L_{d_{3}}\right) $. The parameters $\theta_{f_{1}} \left( \theta_{f_{2}}, \theta_{f_{1, 2}}\right) $ before GRL are updated in the direction of maximizing $L_{d_{1}} \left( L_{d_{2}}, L_{d_{3}}\right) $, and the parameter $\theta_{d_{1}} \left( \theta_{d_{2}}, \theta_{d_{3}}\right)$ after GRL are updated in the direction of minimizing $L_{d_{1}} \left( L_{d_{2}}, L_{d_{3}}\right) $. For the problem of Eq.(\ref{key21}), we use Adam optimizer to update parameters $\theta_{f_{1}}, \theta_{f_{2}}, \theta_{c}$ to minimize $L_{c}$.

\section{Experiments and Results}

\subsection{Dataset}

\begin{table}[htbp]
	\small
	\footnotesize
	\centering
	\caption{Summary of participant demographics, MMSE and MOCA.}
	\resizebox{0.47\textwidth}{!}{
		\begin{tabular}{|c|c|c|c|c|}
			\hline
			\multicolumn{1}{|c|}{}  & Gender    & Age        & MMSE       & MOCA       \\ \hline
			aMCI             & 5 M / 5 F & 65.70$\pm$5.10 & 25.20$\pm$3.26 & 20.10$\pm$3.35 \\ \hline
			HC               & 4 M / 5 F & 68.00$\pm$3.24 & 28.00$\pm$1.22 & 25.00$\pm$2.06 \\ \hline
			\multicolumn{1}{|c|}{$p$} & 0.809     & 0.263      & $0.027^{*}$     & $0.001^{**}$    \\ \hline
			\multicolumn{5}{l}{$^{*}p<0.05\ ^{**} p<0.01$}
		\end{tabular}
	}
	
	\label{tab1}
\end{table}
\begin{table*}[t]
	\small
	\centering
	\caption{Experimental results. \textit{Acc} is the accuracy rate, \textit{ACP} is the average confidence probability, \textit{Sen} is the sensitivity, and \textit{Spe} is the specificity. Resnet-18, Densenet-121, VGG-11-bn all have been pre-trained on ImageNet. 
	}
	\begin{tabular}{c|c|c|c|c|c|c|c|c|c|c|c|c}
		\hline
		\hline
		\multirow{2}{*}{\textbf{Method}} & \multicolumn{4}{c|}{\textbf{DMS Dataset}}                                   & \multicolumn{4}{c|}{\textbf{REST Dataset}}                                   & \multicolumn{4}{c}{\textbf{STROOP Dataset}}                                \\ \cline{2-13} 
		& Acc            & ACP            & Sen            & Spec           & Acc            & ACP            & Sen             & Spe            & Acc            & ACP            & Sen            & Spe            \\ \hline
		KNN \cite{altman1992knn}     & 73.68          & 75.48          & 80.00          & 66.67          & 42.11          & 50.49          & 80.00           & 0.00           & 47.37          & 45.99          & 30.00          & 66.67          \\
		LR classifier \cite{2008LIBLINEAR}     & 78.95          & 75.92          & 70.00          & 88.89          & 57.89          & 56.45          & 90.00           & 22.22          & 42.11          & 43.19          & 20.00          & 66.67          \\
		SVM \cite{cortes1995support-vector}                    & 84.21          & 78.45          & \textbf{90.00} & 77.78          & 57.89          & 56.59          & 90.00           & 22.22          & 68.42          & 
		61.57 & \textbf{80.00} & 55.56          \\
		Random Forest \cite{598994}          & 84.21          & 79.14          & 80.00          & \textbf{89.00} & 52.63          & 42.59          & 90.00           & 11.00          & 57.89          & 59.01          & 60.00          & 55.56          \\
		Resnet-18 \cite{he2016deep}              & 84.21          & 78.28          & 80.00          & 88.89          & 57.89          & 52.47          & \textbf{100.00} & 11.11          & 63.16          & 53.13          & 60.00          & 66.67          \\
		Densenet-121 \cite{huang2017densely}             & 78.95          & 69.71          & 70.00          & 88.89          & 57.89          & 59.83          & 70.00           & 44.44          & 68.42           & \textbf{63.82}          & 70.00           & 66.67              \\
		VGG-11-bn \cite{simonyan2014very}                   & 84.21          & 73.08          & 80.00          & 88.89          & 57.89          & 54.61          & 70.00           & 44.44          & 57.89          & 55.43          & 40.00          & \textbf{77.78} \\ \hline
		GF-DANN(ours)                 & \textbf{89.47} & \textbf{83.86} & \textbf{90.00} & \textbf{89.00} & \textbf{78.95} & \textbf{60.37} & 90.00           & \textbf{66.67} & \textbf{78.95} & 61.03          & \textbf{80.00} & \textbf{77.78} \\ \hline \hline
	\end{tabular}
	\label{tab3}
\end{table*}

\begin{figure}[tbp]
	\centering
	\includegraphics[width=1\linewidth]{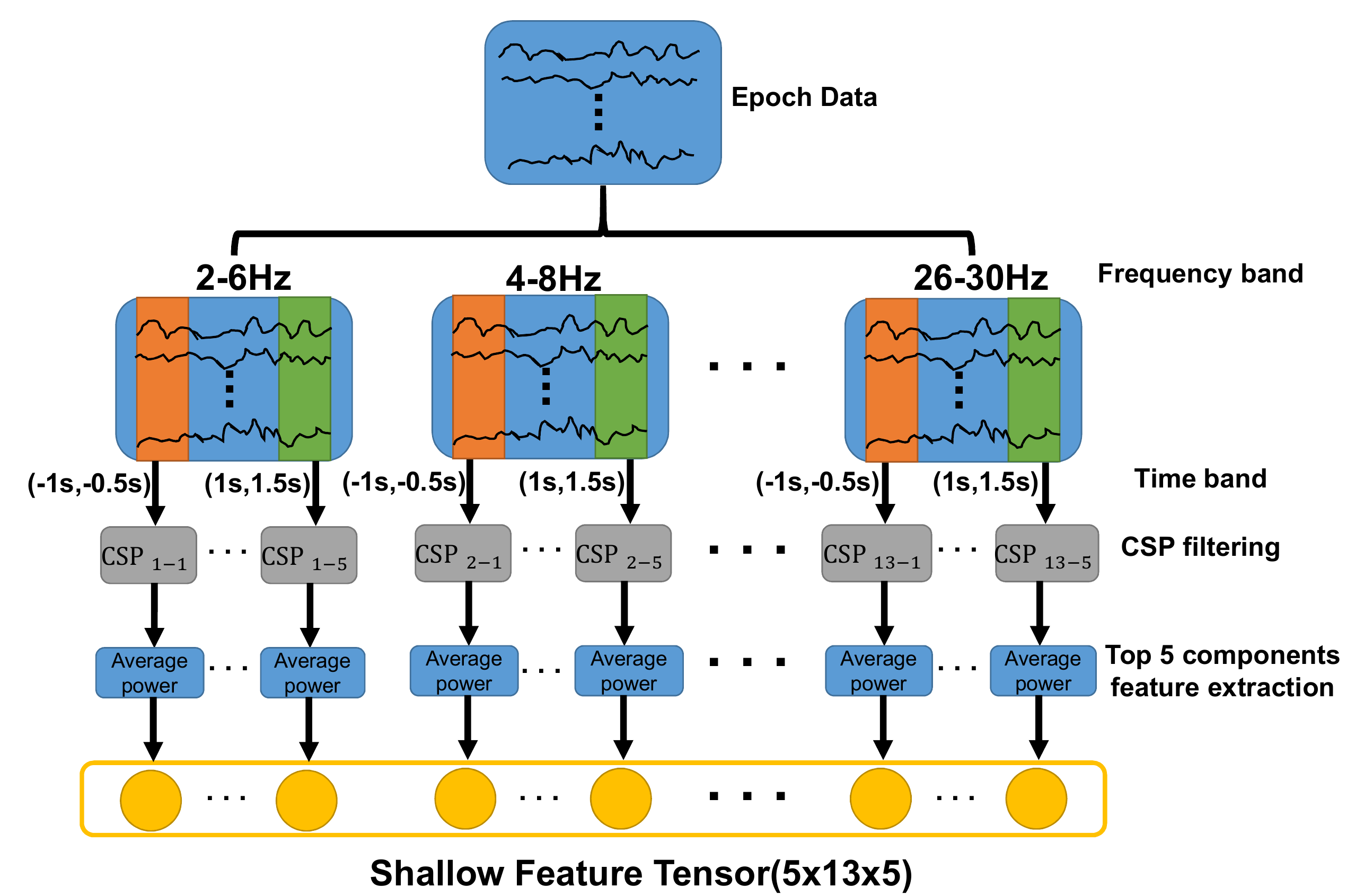}
	\caption{Data preprocess on DMS dataset. }
	\label{fig:preprocess}
\end{figure}

EEG recordings were obtained at the First People’s Hospital of Foshan City, Foshan, China. 
It includes ten patients with aMCI and nine healthy people. The diagnosis results are given by professional doctors based on biomarkers and scale scores. Summary of participant demographics, mini-mental state examination (MMSE), and montreal cognitive assessment (MOCA) are shown in Table \ref{tab1}. There is no significant difference in age ($p=0.263$) and gender ($p=0.809$) distribution between the two groups, and there is a significant difference in the MMSE ($p=0.027$) and MOCA ($p=0.001$) scores. EEG data were collected using Wearable Sensing's DSI-24 EEG device, with a sampling frequency of 300 Hz. Twenty-one sensors positioned according to the 10-20 international system. These electrodes used Pz as the reference electrode during the recording process, so the data of twenty electrodes were finally obtained.

In this study, we collected three types of data by EEG equipment:
(1) DMS \cite{Juan2017A} Dataset: Each subject performed 150 trials. We extract the epoch data corresponding to 2.5s (-1s, 1.5s) at the time of each trial start. The data set size is 19 subjects and 2850 samples.
(2) REST Dataset: participants were asked to close their eyes and stay awake for 2 minutes to obtain resting data. The data set is obtained by extract epoch data at 2s intervals. The data set size is 19 subjects and 1140 samples.
(3) STROOP \cite{stroop1992studies} Dataset: Each subject needs to make 80 judgments, we extract the epoch data corresponding to 3s (-1s, 2s) at the time of each trial start. The data set size is 19 subjects and 1520 samples.
The detailed data collection method can be viewed in the attached video.

\begin{figure*}[t]
	\centering
	\subfigure[$X_{S}^{c_{1}}$ input to GF-DANN]{%
		\includegraphics[width=0.23\linewidth]{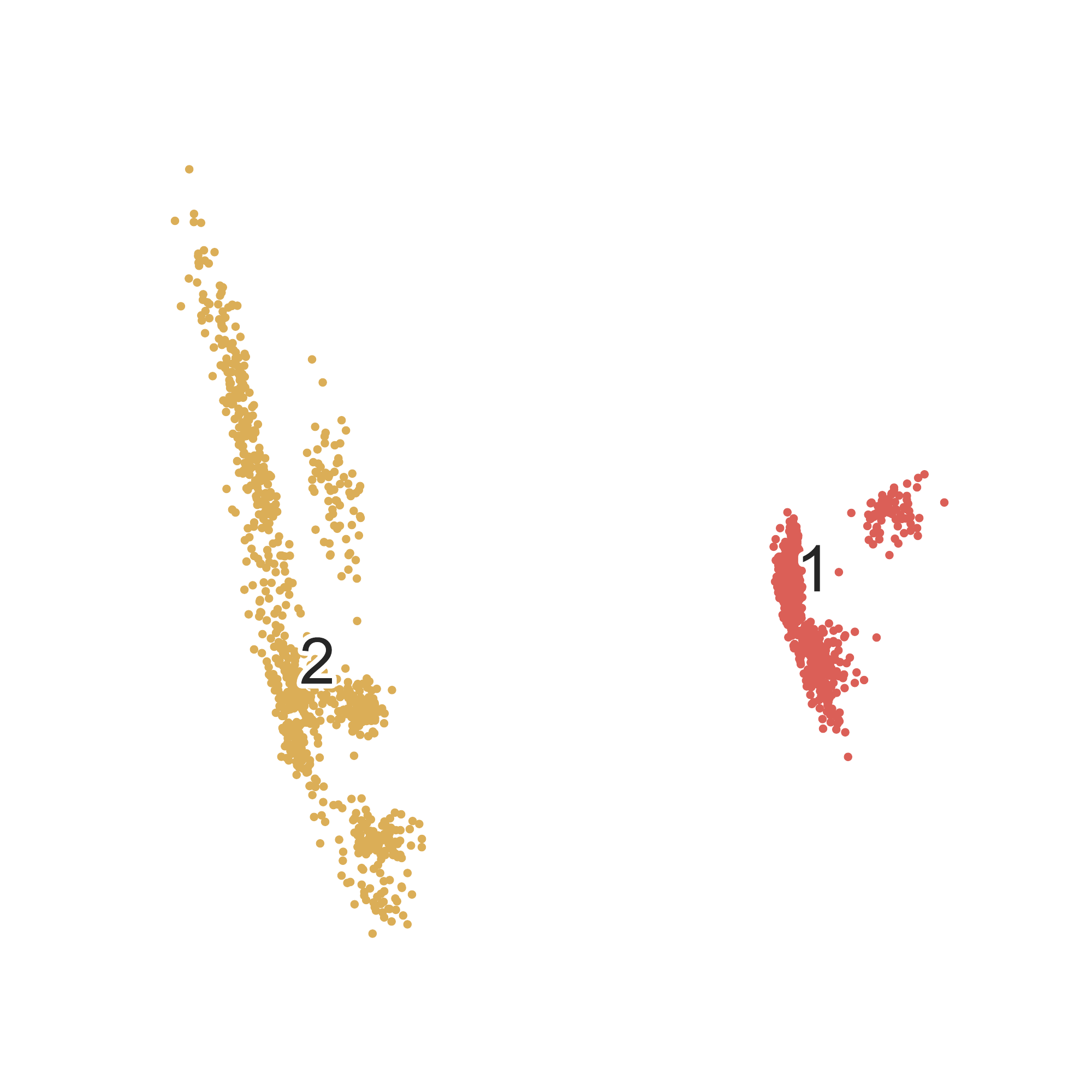}
		\label{fig:x1 input GF-DANN}}
	\subfigure[$X_{S}^{c_{2}}$ input to GF-DANN]{%
		\includegraphics[width=0.23\linewidth]{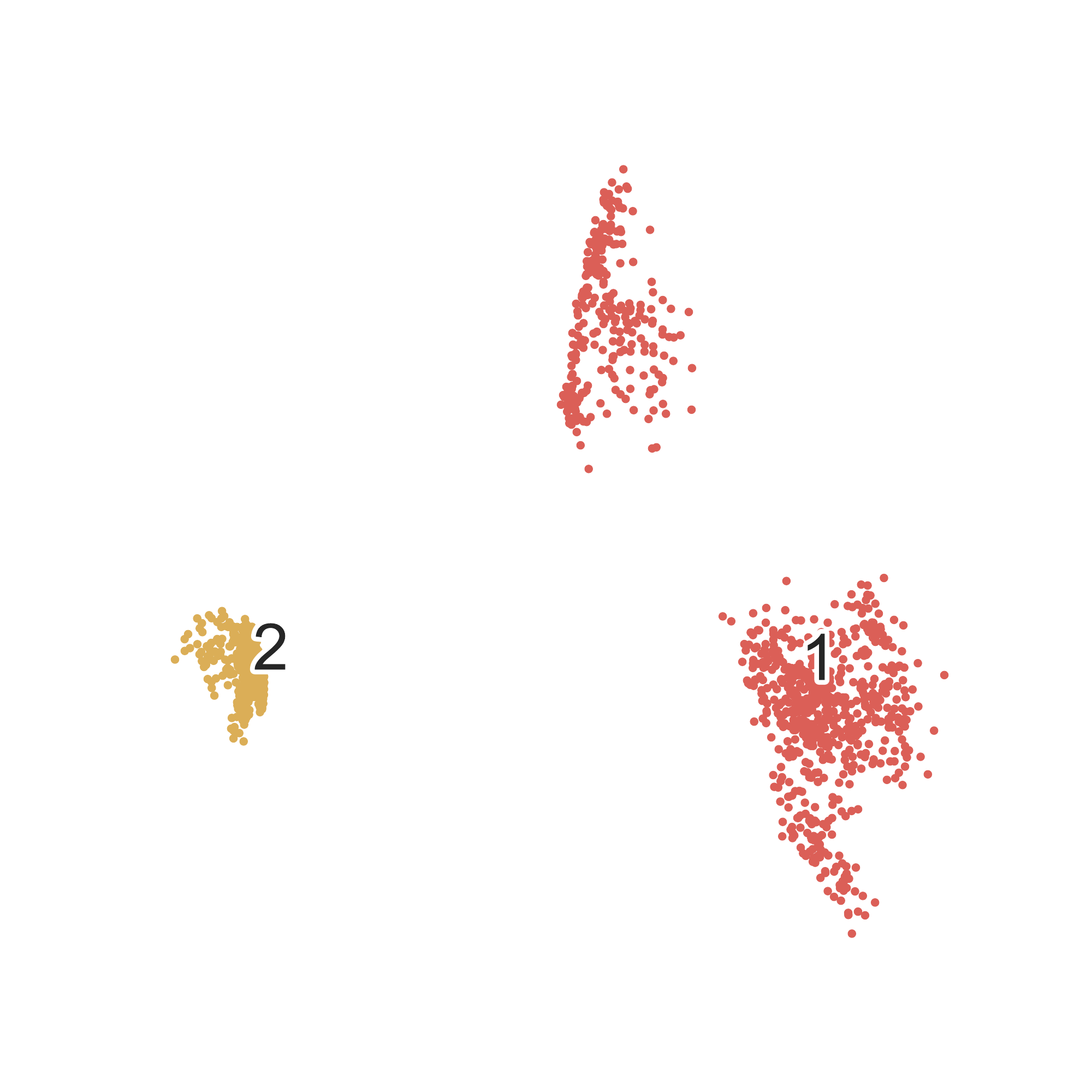}
		\label{fig:x2 input GF-DANN}}
	\subfigure[$X_{S}^{c_{1}}$ input to BaseNet-1]{%
		\includegraphics[width=0.23\linewidth]{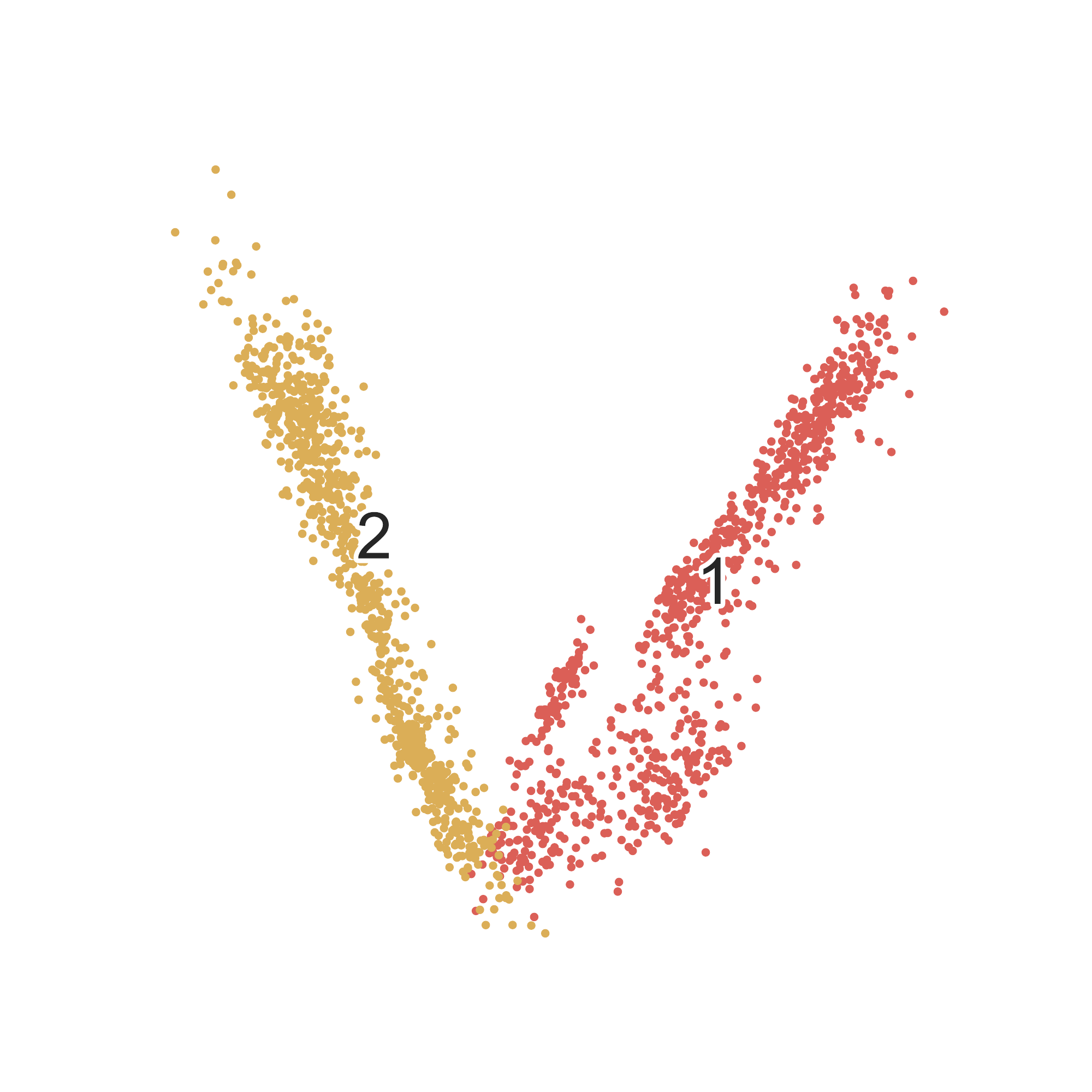}
		\label{fig:x1 input BaseNet-1}}
	\subfigure[$X_{S}^{c_{2}}$ input to BaseNet-1]{%
		\includegraphics[width=0.23\linewidth]{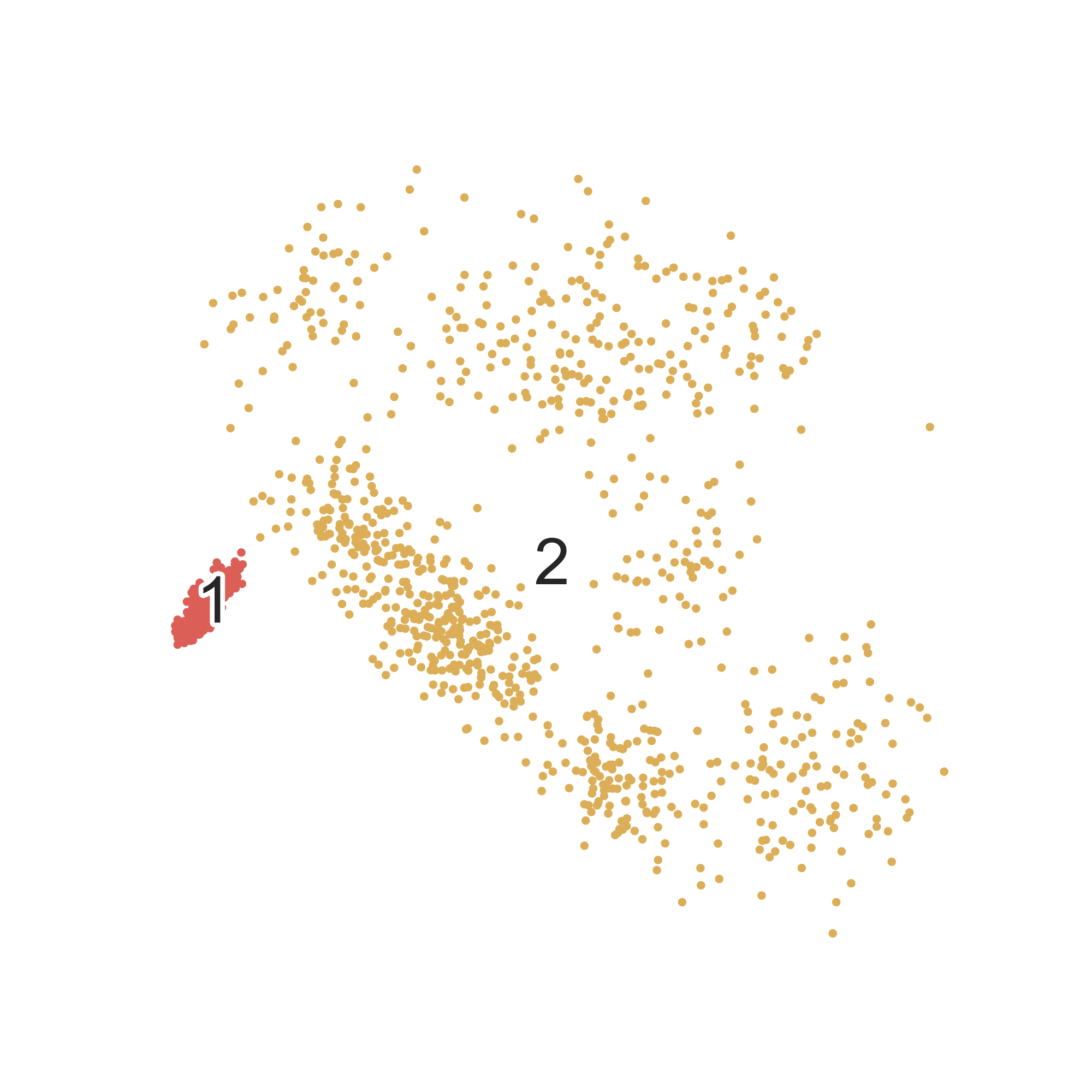}
		\label{fig:x2 input BaseNet-1}}
	\caption{Comparison of vectors obtained by inputting different categories of data for two feature extractors. All points in the figure are the feature vectors obtained by the corresponding input samples through the feature extractor (for visualization, the feature vectors are reduced to two dimensions by PCA), where the red dots and the yellow dots represent the feature vectors extracted by feature extractors 1 and 2, respectively.}
	\label{fig:Feature vector visualization}
\end{figure*}

\subsection{Shallow Feature Extraction}\label{section:Shallow Feature Extraction}
The collected data firstly use the band-pass filtering of 0.5-70 Hz and then apply the notch filtering of 48-52 Hz to eliminate power line interference. We select the epochs with correct responses by subjects and remove the bad epochs by manual inspection. ICA algorithm is used to remove artifacts such as electrooculograms and abnormal movements. Taking the DMS data set as an example, the details of shallow feature extraction are as follows:

As shown in Figure \ref{fig:preprocess}, the epoch data is band-pass filtered with a window length of 4 Hz and a step size of 2 Hz to obtain data in 13 frequency bands among 2-30 Hz. For each frequency band, the data is divided into 5-time bands (window length 0.5s, step size 0.5s). We use the MNE \cite{Gramfort2013MNE} toolkit to design CSP \cite{Blankertz2008Optimizing,Koles1990Spatial} filters for each frequency band and time band data. A total of 65 CSP filters are designed for the DMS data set. 
Note that in the experiment, we use leave-one-group-out cross-validation. In each fold, the corresponding train data is used to design the CSP filter, and then the train and test data are filtered by the designed CSP filter. Therefore, the design of the CSP filter does not cause leakage of test data. Finally, the first five components after filtering are selected, and the average power is obtained for each component to obtain the shallow feature tensor.

\subsection{Implementation Details}
We use leave-one-group-out cross-validation to evaluate model performance. Each time, select samples of one subject (1/19) as the test set and the remaining subject's (18/19) data as the train set. According to the voting diagnosis framework, in the testing stage, the corresponding sample-level classification results $\tilde{y}_{i} \in \{0,1\},i=1,\cdots,k$ of the test subject are obtained through GF-DANN. Subject-level diagnosis results $\tilde{y}_{\text{subject}}$ were generated by voting on the classification results of these sample levels. 
The confidence probability $P_{i}$ of a single subject $i$ is defined as follows:
\begin{footnotesize} 
	\begin{equation}\label{key22}
	P_{i} =\frac{N_{\text{correct}}(i)}{N(i)}
	\end{equation}
\end{footnotesize}

\noindent where $N_{\text{correct}}(i)$ represents the number of correctly classified samples of subject $i$, $N(i)$ represents the number of all samples of subject $i$.

Define the average confidence probability (ACP) as follows:
\begin{footnotesize} 
	\begin{equation}\label{key23}
	A C P=\frac{1}{N_{\text{subject}}} \sum_{i=1}^{N_{\text{subject}}} P_{i}
	\end{equation}
\end{footnotesize} 

\noindent where $N_{\text{subject}}$ represents the number of all subjects.

In this paper, the classification accuracy, average confidence probability (ACP), sensitivity, and specificity are used as evaluation indicators.
Traditional machine learning and classic deep learning methods are used to compare with GF-DANN. The machine learning method is based on scikit-learn and performs a parameter search to obtain the best performing model. Deep learning methods and GF-DANN is constructed based on PyTorch and trained using Tesla V100-SXM2-32GB, using Adam as the optimizer, focal loss \cite{Lin2017Focal} as the loss function. A learning rate decay strategy is used during the training process. 
Each model is tested on the corresponding test set after training 20 epochs.

\begin{table}[t]
	\small
	\centering
	\caption{Ablation study for GFE and DBDA on DMS dataset. 
	}
	\resizebox{0.47\textwidth}{!}{
		\begin{tabular}{c|c|c|c|c|c|c}
			\hline
			\hline
			\textbf{Method}   & \textbf{GFE} & \textbf{DBDA} & \textbf{Acc} & \textbf{ACP} & \textbf{Sen} & \textbf{Spe} \\ \hline
			BaseNet-1 &             &             & 78.95             & 77.57        & 80.00                & 78.00                \\
			BaseNet-2 & $\surd$           &             & 89.47             & 78.41        & 90.00                & 89.00                \\
			BaseNet-3 &             & $\surd$           & 89.47             & 79.93        & 90.00                & 89.00                \\
			GF-DANN   & $\surd$           & $\surd$           & 89.47             & 83.86        & 90.00                & 89.00                \\ \hline \hline
		\end{tabular}
	}
	\label{tab2}
\end{table}
\subsection{Ablation Study}

GFE is used to make the feature extractor learn the common features of the individuals in the group. DBDA is used to reduce the feature distribution gap between the source domain and the target domain. The ablation experiment is used to demonstrate the performance of the discriminator. The experimental results are shown in Table \ref{tab2}. 

GF-DANN without GFE and DBDA is used as the base network. When using GFE or DBDA, the network outperforms the base network 10.52\%, 10\%, and 11\% by accuracy, sensitivity, and specificity. The effect of using both two types of discriminators is mainly reflected in the improvement of the ACP. The network applying DBDA outperforms the base network by 2.36\% ACP. The ACP increases by 0.84\% due to the use of GFE. The network applyling both GFE and DBDA achieves 83.86\% ACP, which outperforms the base network by 6.29\% ACP. These improvements indicate that GFE and DBDA can significantly improve classification accuracy.

\subsection{Results on Three Dataset}

GF-DANN achieved the best accuracy rates on all three data sets (Table \ref{tab3}), increasing by 5.26\%, 21.06\%, and 10.53\% respectively from the second place method. It is worth mentioning that in the DMS dataset, the Acc, ACP, Sen, and Spe of our method are 89.47\%, 83.86\%, 90\%, and 89\%, respectively, which demonstrates a result with potential for clinical application.
All methods have achieved higher scores in the DMS data set than the REST and STROOP data sets, which shows that the DMS acquisition paradigm is more suitable for the diagnosis of aMCI.  
The main clinical manifestation of aMCI is memory impairment, which is more in line with the investigation of working memory ability that DMS focuses on, which may be the reason for the better diagnosis of DMS data sets.

\subsection{Feature Vector Visualization}

To confirm whether the two feature extractors of GF-DANN have learned the corresponding group features, we conducted a feature vector visualization experiment.
Input $X_{S}^{c_{1}}$ or $X_{S}^{c_{2}}$ into the trained GF-DANN or BaseNet-1 to obtain the feature vector sets F1 and F2 of the corresponding class samples through the two feature extractors. After that, PCA dimension reduction is performed on the feature vectors for two-dimensional visualization.
Figure \ref{fig:Feature vector visualization} shows the feature vector visualization for four configurations. All points in the figure are the feature vectors obtained by the corresponding input samples through the feature extractor, where the red dots and the yellow dots represent the feature vectors extracted by feature extractors 1 and 2, respectively. For example, Figure \ref{fig:x1 input GF-DANN} shows the two-dimensional visualization of the corresponding two feature vector sets obtained by the trained feature extractor extract class 1 data $X_{S}^{c_{1}}$. It represents the distribution of the class 1 data after passing through two feature extractors.

Comparing Figure \ref{fig:x1 input GF-DANN} and \ref{fig:x2 input GF-DANN}, it can be seen that when the feature extractor extracts the corresponding category data, the resulting feature vector distribution is more concentrated. In contrast, when extracting other categories of data, the feature vector distribution has a dispersed distribution. This analysis 
demonstrates that
the feature extractor of GF-DANN has learned the common features of the corresponding categories in a targeted manner, thus obtaining a more compact distribution. However, similar phenomena cannot be observed in Figure \ref{fig:x1 input BaseNet-1} and \ref{fig:x2 input BaseNet-1}.

Compared with BaseNet-1, the distance among the feature vector sets learned by the two feature extractors of GF-DANN is more considerable. This shows that the two feature extractors learn features with more significant differences, which helps obtain more features for classification.

\section{Conclusion}
This paper proposes a novel neural network and introduces the idea of adversarial learning to solve the problem of huge differences in individual data in aMCI diagnosis, which has achieved better results than classical machine learning and deep learning methods during three types of EEG data sets. 
We believe the proposed method takes a significant step in constructing an EEG based medical diagnostic robot system.



%

\section*{ACKNOWLEDGMENT}

This work was supported in part by the National Key R\&D Program of China (Grant 2018YFC2001700), National Natural Science Foundation of China (Grants 61720106012, U1913601, 62073319, 62003343, U20A20224), Beijing Natural Science Foundation (Grant L172050), the Youth Innovation Promotion Association of CAS under Grant 2020140, and by the Strategic Priority Research Program of Chinese Academy of Science (Grant XDB32040000).


\bibliographystyle{IEEEtran}
\bibliography{root.bib}

\end{document}